\documentclass[10pt, conference]{IEEEtran}
\IEEEoverridecommandlockouts
\usepackage{cite}
\usepackage{amsmath,amssymb,amsfonts}
\usepackage{algorithmic}
\usepackage{graphicx}
\usepackage{textcomp}
\usepackage{xcolor}
\usepackage[normalem]{ulem}

\usepackage{url}
\usepackage{multirow}
\usepackage{float}

\def\BibTeX{{\rm B\kern-.05em{\sc i\kern-.025em b}\kern-.08em
    T\kern-.1667em\lower.7ex\hbox{E}\kern-.125emX}}

\begin{document}

\title{Dynamic Transformer for Efficient Machine Translation on Embedded Devices}

\author{\IEEEauthorblockN{Hishan Parry, Lei Xun, Amin Sabet, Jia Bi, Jonathon Hare, Geoff V. Merrett}
\IEEEauthorblockA{\textit{School of Electronics and Computer Science} \\
\textit{University of Southampton}\\
Southampton, UK \\
{\{hmrp1r17, lx2u16, ms4r18, J.Bi, jsh2, g.merrett\}}@southampton.ac.uk}

}


\maketitle

\IEEEpubidadjcol

\begin{abstract}
The Transformer architecture is widely used for machine translation tasks. However, its resource-intensive nature makes it challenging to implement on constrained embedded devices, particularly where available hardware resources can vary at run-time. We propose a dynamic machine translation model that scales the Transformer architecture based on the available resources at any particular time. The proposed approach, `Dynamic-HAT', uses a HAT SuperTransformer as the backbone to search for SubTransformers with different accuracy-latency trade-offs at design time. The optimal SubTransformers are sampled from the SuperTransformer at run-time, depending on latency constraints. The Dynamic-HAT is tested on the Jetson Nano and the approach uses inherited SubTransformers sampled directly from the SuperTransformer with a switching time of \textless 1s. Using inherited SubTransformers results in a BLEU score loss of $\leqslant$ 1.5\% because the SubTransformer configuration is not retrained from scratch after sampling. However, to recover this loss in performance, the dimensions of the design space can be reduced to tailor it to a family of target hardware. The new reduced design space results in a BLEU score increase of approximately 1\% for sub-optimal models from the original design space, with a wide range for performance scaling between 0.356s - 1.526s for the GPU and 2.9s - 7.31s for the CPU.
\\
\end{abstract}


\section{Introduction}
\begin{figure}[]
\centerline{\includegraphics[width=9.2cm]{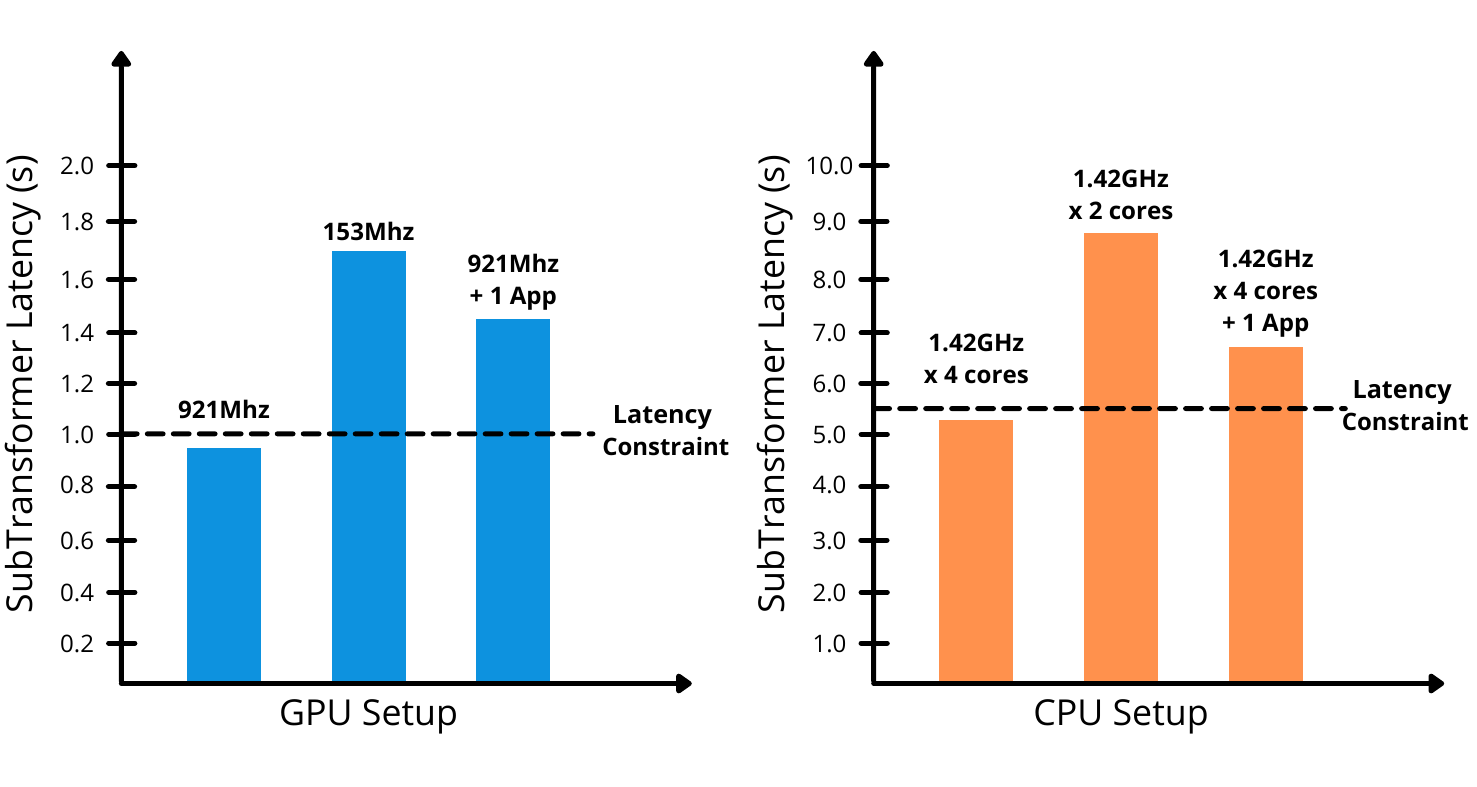}}
\caption{Experimental results to showing how latency constraints set during design time optimization are not met at run-time for a Transformer model on the Jetson Nano, due to a varying availability of computing resources or power.}
\label{motivation_fig}
\end{figure}

Machine translation is a fast-growing application of Natural Language Processing (NLP) tasks on embedded devices. However, the most widely used machine learning architecture for NLP, the Transformer \cite{transformer_vanilla}, is computationally expensive and hence challenging to deploy on embedded devices with limited computational and energy resources. Various approaches have been proposed to address the computational and energy demands of Transformers on embedded devices, such as the Lite Transformer \cite{lite_transformer}, which combines convolution with the standard attention mechanism of the Transformer followed by pruning, allowing better performance in embedded NLP applications. These approaches, which are static DNNs, provide an optimal Transformer architecture for the target application performance requirements based on the measurement
of fixed hardware resources. However, embedded devices often run many applications on several heterogeneous cores, and the resources a Transformer was optimized for may not be available at run-time \cite{optimising_resource}. Therefore, the performance requirements of the application can be violated. Figure \ref{motivation_fig} illustrates how the desired latency constraint set for a Transformer model can be violated at run-time on the Jetson Nano. For example, on both the GPU and CPU setups, the latency constraint holds when all resources  can be dedicated to the Transformer model at full power. When the Jetson is in low-power mode, the reduced frequency of the GPU or the reduced number of CPU cores are not able to meet the required latency. The latency constraint is also not met when the Jetson is at full power, but is dedicating resources to other applications. The frequency and number of cores used may vary at run-time due to power constraints or other processes running simultaneously and using up resources.

Given that hardware resource availability can change dynamically at run-time, dynamic DNNs \cite{dynabert}\cite{scalable_transformer} have
been proposed to address this issue. A dynamic DNN
contains various sub-networks, each with its accuracy and resource requirements. Given varying available hardware resources and application requirements, different subnetworks can be selected. Previous approaches contain an additional training pipeline to generate these subnetworks. However, state-of-the-art NAS approaches \cite{hat} already have an efficient backbone supernetwork and hence, the additional training pipeline can be skipped.

\begin{figure*}[h]
\centerline{\includegraphics[width=18.8cm, height=9.1cm]{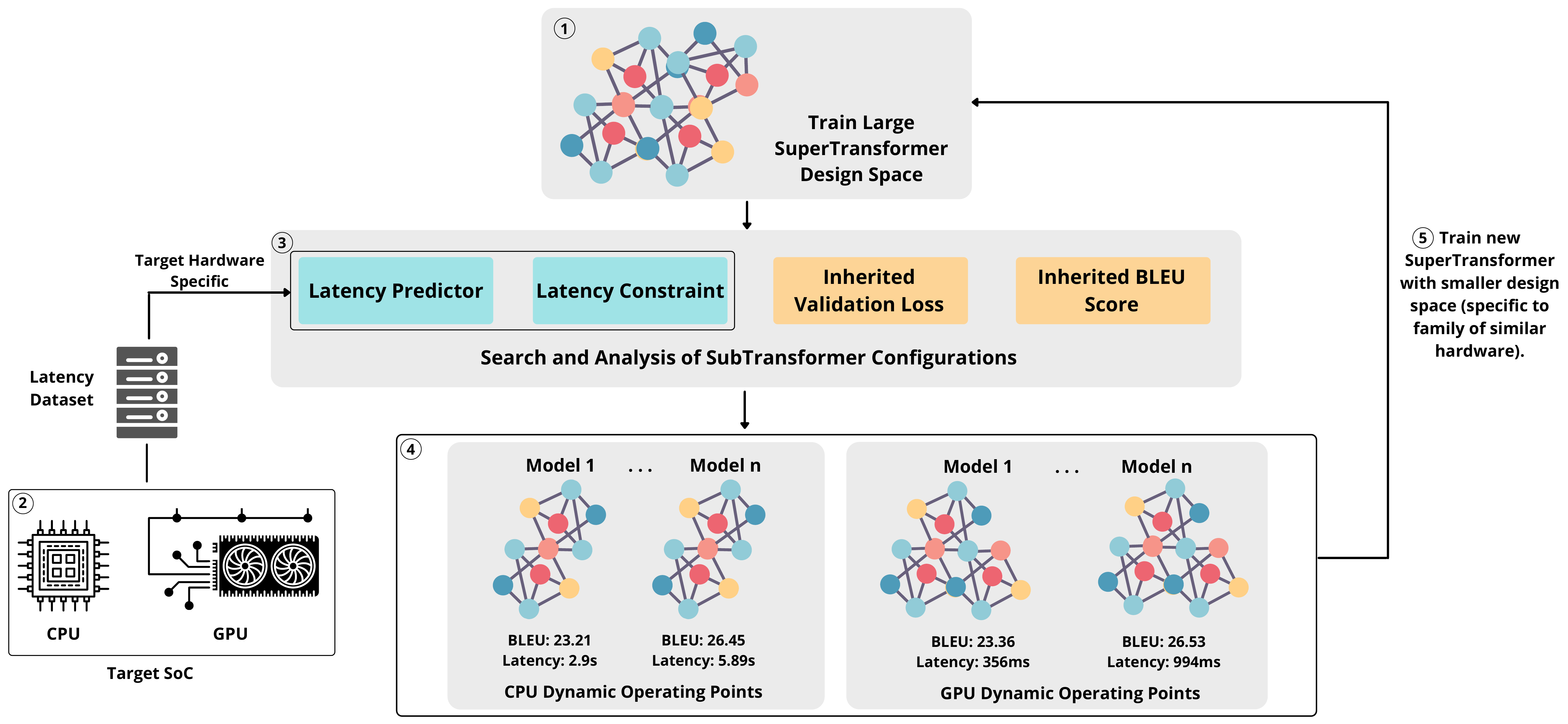}}
\caption{An overview of the steps in the proposed Dynamic-HAT: (1) A SuperTransformer with a large design space is first trained. (2) Latency datasets are collected from the target hardware. (3) Optimal SubTransformers are searched for the target hardware based on the provided latency constraints. The SubTransformer configurations are analyzed to find the dimensions of the design space preferred by the target hardware (or family of target hardware). (4) These searched SubTransformer configurations are used by the Dynamic-HAT as operating points to meet latency constraints on time-varying hardware resources. (5) The last step is to train a new SuperTransformer with a reduced design space based on the target hardware. This attempts to mitigate the performance loss arising from the use of inherited SubTransformers.}
\label{dhat_figure}
\end{figure*}

This paper proposes Dynamic-HAT, a dynamic Transformer that can change the complexity of its architecture at run-time. The approach uses Hardware-Aware Transformers (HAT) \cite{hat} as the backbone. First, a SuperTransformer is trained once. The SuperTransformer has a large design space containing $10^{15}$ SubTransformers which encompass optimal model architectures for various accuracy-latency trade-offs. The evolutionary search in HAT to find an optimal SubTransformer to fit a given latency constraint is too expensive to perform at run-time. Hence, we pre-select a small set of optimal SubTransformer configurations for fast switching at run-time. This approach is flexible as it allows changing several hidden dimensions of the Transformer architecture through the HAT backbone with no need for retraining before inference, unlike previous dynamic DNN approaches. These Inherited SubTransformers, which are sampled directly from the SuperTransformer, closely approximates a similar configuration fully trained from scratch with a slight decrease in performance. We also explore different techniques to reduce the performance loss when using Inherited SubTransformers. Sampling inherited SubTransformers allows scaling the Transformer model at run-time to fit changing latency constraints, meaning an optimal Transformer model can always be found for different availability of resources on the same device. The contributions of this paper are:

\begin{itemize}
  \item Dynamic Transformer architecture using HAT as the backbone of a SuperTransformer, allows SubTransformer configurations to be sampled at run-time without additional training.
  \item A run-time approach to dynamically switch between
Transformer architectures with switching time \textless 1s. Dynamic model switching allows the device to meet time-varying hardware resource availability, and maintain performance constraints of machine translation applications.
  \item Dynamic-HAT reduces the SuperTransformer design space for a set of similar target hardware devices, resulting in a BLEU score increase of up to 1\%. We demonstrate Dynamic-HAT using the Jetson Nano, with a wide range for performance scaling between 0.356s - 1.526s for the GPU and 2.9s - 7.31s for the CPU.
\end{itemize}

\section{Related Work}
Several methods have been developed to counter the computational complexity of the Transformer model for efficient NLP, which can be divided into static and dynamic approaches. 

\subsection{Static Approaches}
Although standard machine translation models such as the original Transformer \cite{transformer_vanilla} and BERT \cite{bert} are computation and resource-intensive, there exist off-line model compression approaches that allow deployment on embedded hardware platforms \cite{deep_compression} \cite{netadapt}. A popular method of static model compression is pruning, as proposed by Zhu et al. \cite{pruning}, where sparsity is introduced to deep neural networks in an attempt to reduce the number of computations and make a smaller model. Another static approach to make machine learning models more efficient on resource-limited hardware is quantization, following Jacob et al. \cite{quantization}, which lowers data precision to reduce memory requirements and computational cost. However, most of these works only compress the model to a fixed size (static), such as the Lite Transformer \cite{lite_transformer}, so the proposed model may not be optimal at run-time when resource constraints change.

\subsection{Dynamic Approaches}
An example of a dynamic model for NLP tasks is DynaBERT \cite{dynabert}, a BERT model which can adjust the model size and latency by changing the width and depth of the network. Another approach to dynamic neural machine translation is the scalable transformer \cite{scalable_transformer}, which can change the feature dimensions for all sub-layers in the encoder/decoder. Pruning can also be used for dynamic implementations, such as dynamic sparsity neural networks, which have a set number of model sparsity levels that can be changed at run-time \cite{dynamic_sparsity}.

Compared to previous dynamic approaches, the proposed Dynamic-HAT is more flexiable due to the use of a HAT backbone \cite{hat}. HAT is more similar to the original Transformer, but with a larger number of elastic hidden dimensions. HAT also has the benefit of arbitrary encoder-decoder attention and flexible decoder layers. HAT addresses the environmental concerns of Transformers by training a SuperTransformer (encompassing a large design space) only once. This SuperTransformer is used to sample SubTransformers for different target hardware under varying latency constraints. The design space encompasses about $10^{15}$ SubTransformer configurations resulting in a wide range of model sizes and latencies, which makes it possible to find a SubTransformer model for a range of hardware platforms spanning from powerful computers with NVIDIA GPU’s to tiny IoT devices with ARM CPUs. Instead of just targeting different hardware platforms, our approach moves one step further by using a library of pre-selected optimal SubTransformers to target dynamically available hardware resources at run-time. The small library also enables our approach to have real-time model switching capability, which is not possible using a static HAT model.

Dynamic-OFA \cite{dynamic_ofa} is an image classification dynamic DNN based on the Once-for-All (OFA) approach \cite{ofa}, where a large model is first trained from scratch. Several optimal sub-networks on the Pareto front can be pre-sampled from this large model at design time, and adapted to best match the hardware constraints during run-time. Our Dynamic-HAT approach takes inspiration from Dynamic-OFA, whereas our application is NLP.

\section{Dynamic-HAT}
This section introduces our proposed dynamic machine translation model that is built based on HAT \cite{hat}. Figure \ref{dhat_figure} shows an overview of the proposed Dynamic-HAT.  A SuperTransformer with a large design space is first trained from scratch. Different SubTransformer configurations can be sampled from the SuperTransformer and used for inference during run-time, depending on resource constraints, which allows the Transformer model to be trained only once but used for a variety of constraints.

\begin{table}[]
\centering
\caption {Performance Comparison between Inherited SubTransformers and SubTransformers trained from scratch. The results are obtained by running HAT architecture on the Jetson Nano GPU and comparing directly sampled SubTransformers to their retrained counterparts.}
\begin{tabular}{c|c|c}
\hline
\textbf{Latency)} & \textbf{Inherited} & \textbf{From-Scratch} \\ 
\textbf{Constraint (ms)} & \textbf{BLEU Score} & \textbf{BLEU Score} \\ 
\hline
500 & 26.17 & 26.33 \\ 
\hline
1000 & 26.67 & 26.84 \\
\hline
1500 & 26.26 & 27.21 \\
\hline
\end{tabular}
\end{table}

\subsection{Proposed Dynamic Model for Machine Translation}

The final step in the HAT \cite{hat} process is to train the searched SubTransformer from scratch before being deployed to the target hardware. Table I shows a performance comparison between sampled Inherited SubTransformers and their counterparts that have been trained from scratch. Retraining before inference increases the performance of the SubTransformer configuration by approximately 1.5\% on average. Inherited SubTransformers perform similar to SubTransformers retrained from scratch before inference. This is because all SubTransformers in the SuperTransformer design space are uniformly trained with shared weights. Hence, Inherited SubTransformers have the same relative performance order as SubTransformers trained from scratch.

Using Inherited SubTransformers would be the more favourable approach when designing a dynamic model based on HAT \cite{hat} as the model does not need to be retrained before inference. Hence, we get a wide variety of model configuration choices at run-time, with only one training cost. Another option, training the searched SubTransformer configurations again from scratch and switching between these trained individual models, although better in performance, is still memory expensive and results in a bigger carbon footprint for training. As illustrated by Figure \ref{dhat_figure}, our approach to Dynamic-HAT contains the following steps:
\begin{enumerate}
  \item A SuperTransformer model for the specific translation task is trained once, with uniformly trained weight-shared SubTransformers (adapted from HAT \cite{hat}).
  \item A latency dataset is created on the target hardware device.
  \item An evolutionary search is performed to find the optimal SubTransformer configuration for various latency constraints supported by the target hardware. The part of the design space used by the target hardware for this latency range is analyzed and recorded.
  \item The searched set of optimal SubTransformers are used as operating points to switch the SubTransformer model when latency constraints change at run-time.
  \item A new SuperTransformer design space can be trained, with reduced dimensions specialized for a product family of similar target hardware, leading to performance improvements in BLEU score.
\end{enumerate}

\subsection{Reducing the Search Space}
This section elaborates on Step 5 of the Dynamic-HAT approach detailed in Figure \ref{dhat_figure}. Our dynamic approach using Inherited SubTransformers, as mentioned previously, results in approximately $\leqslant$ 1.5\% degradation in BLEU score as the model is not retrained before inference. To mitigate this loss in performance, we attempt to reduce the search space of the SuperTransformer to decrease the number of parameters and model configurations. Traditionally, the HAT approach contains a large design space containing SubTransformers suitable for embedded devices and high-end GPUs. Different devices have varying hardware capabilities, and the SubTransformer model searched for each device at a specific latency constraint will be different. However, devices running on similar hardware configurations perform better using the same type of model structure and, thus, specific parts of the SuperTransformer space. For example, when running an evolutionary search, GPUs find wide but shallow models while CPUs find deep but thin models \cite{hat}. Hence, it would be beneficial to reduce the SuperTransformer design space and tailor it to a family of the target hardware, such as ARM CPUs. 

This process has to be done by analyzing the searched HAT models for the target devices, choosing only the parameters in the best-performing configurations, and eliminating the rest. As this procedure would require a new SuperTransformer to be trained from scratch (with fewer parameter choices), training one SuperTransformer for a large group of devices with similar hardware configurations would be the more environmentally friendly approach. As a result, this step remains optional in our dynamic machine translation model, with the additional benefit of improved translation performance.

\section{Experimental Evaluation}

\subsection{Experiment Setups}
\label{setup}
\paragraph{\textbf{Datasets}}
The experiments are conducted on the WMT’14 En-De translation task consisting of 4.5M pairs of training sentences. We use 32K source-target byte pair encoding for the vocabulary that trains on WMT’16, validate on newstest2013, and test on newstest2014.  
\paragraph{\textbf{Evaluation Metrics}}
To evaluate the effectiveness of translated sentences, we use the BLEU score with compound splitting, and case-sensitive tokenization, consistent with \cite{transformer_vanilla,hat}. The BLEU metric evaluates a machine translation by comparing it to a reference human translation, where a score of 100 is a perfect match and 0 is a perfect mismatch \cite{bleu}. The latency of a model is measured by recording the time taken to translate from a source sentence to a target sentence of the same length (30 words). The latency is measured 300 times, removing the fastest and slowest 10\% and taking the average of the other 80\%, consistent with \cite{hat}. The baseline Transformer used for the SuperTransformer in this experiment is the HAT approach \cite{hat}. 
\paragraph{\textbf{Hardware Platform Settings}}
Our experiments are implemented on an NVIDIA Jetson Nano, which is an embedded computer with an ARM A57 CPU and a 128-core Maxwell GPU for inference. As the Jetson Nano is a compact and powerful computer specifically made to run machine learning models within a low power budget of 5-10 Watts, it allows us to evaluate the performance of SubTransformers for a wide range of latency constraints.
\paragraph{\textbf{SuperTransformer Setups}}
Similar to the original Transformer from \cite{transformer_vanilla}, the SuperTransformer in HAT consists of an encoder-decoder structure. Each encoder/decoder consists of multiple layers of stacked identical encoders/decoders. It contains Multi-Head Attention, Feed Forward Network (FFN), and Masked Multi-Head Attention sub-layers with elastic dimensions, which allows changing the SubTransformer configuration. The number of encoder/decoder layers and their input embedding dimensions are also elastic.

Unlike the encoder, the decoder cannot be parallelized due to its auto-regressive nature, which results in the decoder performing a larger number of computations than the encoder.  As the encoder accounts for less than 5\% of the total measured latency of the model,  we are suggested by HAT to fix the number of encoder layers at 6 \cite{hat}. Following the pre-trained SuperTransformer setup from HAT, the design space has the following choices for each parameter:

\begin{itemize}
  \item \{512, 640\} input embedding dimensions for the encoder/decoder
  \item \{1024, 2048, 3072\} number of hidden FFN dimensions
  \item \{4, 8\} number of attention heads in each Multi-Head Attention sub-layer
  \item \{1, 2, 3, 4, 5, 6\} number of decoder layers
  \item \{-1, 1, 2\} for arbitrary encoder-decoder attention: -1 means attending to last one encoder layer, 1 means last two encoder layers, 2 means last three encoder layers.
\end{itemize}

\begin{figure}[!htb]
\centerline{\includegraphics[width=9.25cm]{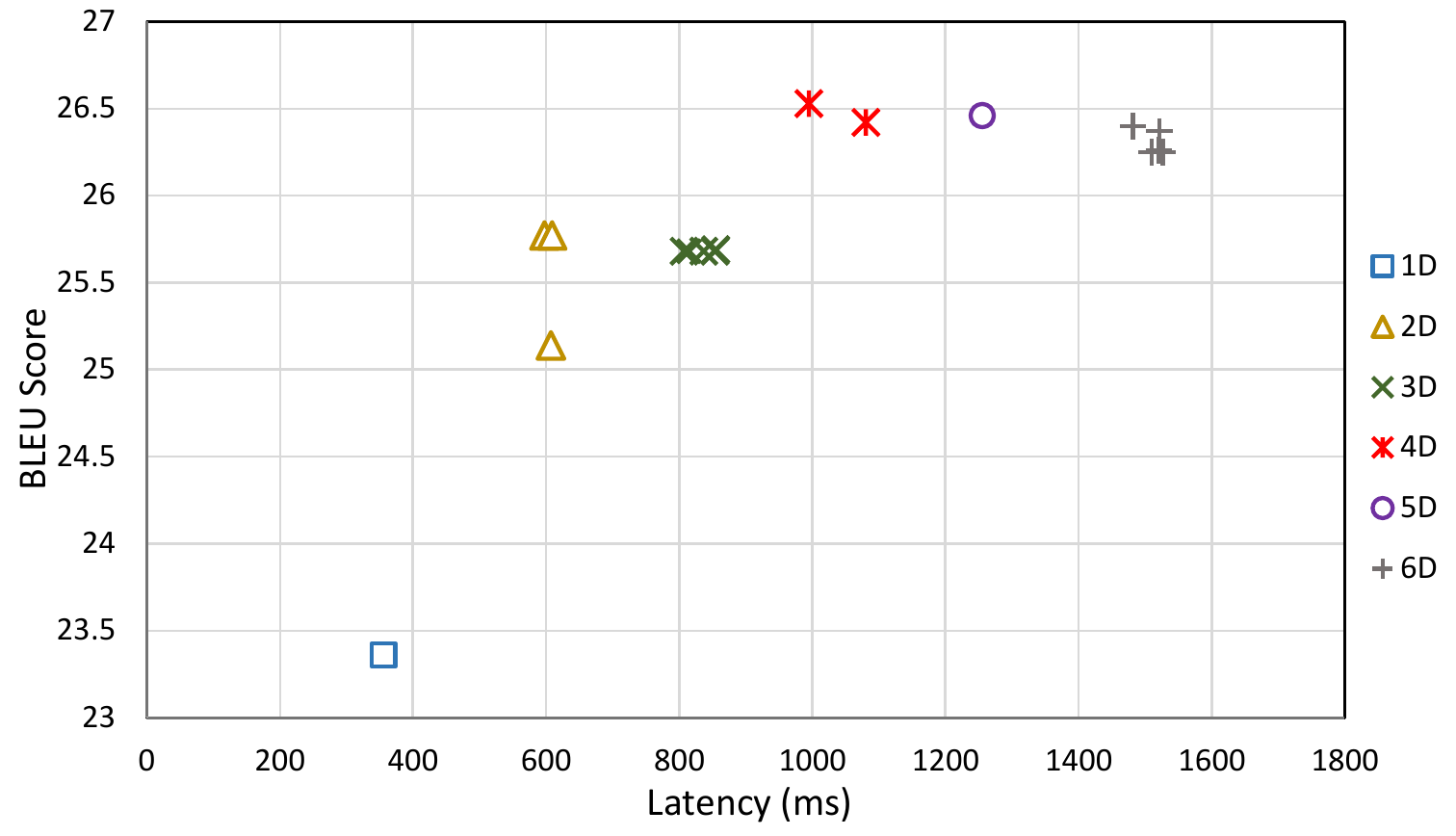}}
\caption{BLEU Score vs experimentally measured Latency on the Jetson Nano GPU, showing how the number of decoder layers affect translation performance of the sampled SubTransformer model.}
\label{fig1}
\end{figure}

\begin{table*}
\centering
\caption{Searched SubTransformer operating points for the Jetson Nano with inherited BLEU scores and validation loss.}
\begin{tabular}{cccc|cccc}
\hline
\multicolumn{4}{c}{\textbf{Jetson   Nano GPU}}
& \multicolumn{4}{c}{\textbf{Jetson   Nano CPU}}                                                 \\
\hline
\textbf{\begin{tabular}[c]{@{}c@{}}Measured\\Latency (ms)\end{tabular}} &
\textbf{\begin{tabular}[c]{@{}c@{}}Inherited \\ BLEU Score\end{tabular}} & \textbf{\begin{tabular}[c]{@{}c@{}}Inherited\\ Validation Loss\end{tabular}} &
\textbf{\begin{tabular}[c]{@{}c@{}}No. Decoder \\ Layers\end{tabular}} &
\textbf{\begin{tabular}[c]{@{}c@{}}Measured\\Latency (s)\end{tabular}} & 
\textbf{\begin{tabular}[c]{@{}c@{}}Inherited   \\ BLEU Score\end{tabular}} & \textbf{\begin{tabular}[c]{@{}c@{}}Inherited \\ Validation Loss\end{tabular}} &
\textbf{\begin{tabular}[c]{@{}c@{}}No. Decoder \\ Layers\end{tabular}} \\
\hline
356.11 & 23.66 & 4.8229  & 1      & 2.90  & 23.21  & 4.9437  & 1  \\
608.95  & 26.23   & 4.3821 & 2    & 3.96 & 25.73  & 4.4163  & 2 \\
854.85  & 26.17   & 4.231 & 3     & 4.97 & 25.63  & 4.2336  & 3  \\
994.96   & 26.67   & 4.155 & 4    & 5.90 & 26.45  & 4.1478  & 5   \\
1255.38  & 26.52  & 4.1177 & 5  & 6.75 & 26.31 & 4.114 & 6  \\
1526.54   & 26.53 & 4.1048 & 6    & 7.31 & 26.33  & 4.1044 & 6 \\
\hline
\end{tabular}
\label{table2}
\end{table*}

\subsection{Linking Decoder Layers to Performance and Latency}
Figure \ref{fig1} shows the relationship between the BLEU score and measured latency for several operating points on the Jetson Nano GPU. It is observed that the performance of a searched SubTransformer configuration does not increase linearly with latency constraints. The actual performance of the model depends significantly on the number of decoder layers being used. The ‘gaps’ in the graph between data points occur when the number of decoder layers change. Adding one decoder layer to the SubTransformer configuration increases its measured latency by around 300 ms on the GPU and around 1000ms on the CPU.

A higher number of decoder layers does not necessarily lead to better performance in the model, since it greatly depends on the hardware that is running the model. As mentioned before, GPUs perform better with wide but shallow models while CPUs prefer deep but thin models \cite{hat}. As GPUs are less sensitive to changes in hidden dimensions and embedded dimensions, while CPUs are very sensitive,  a GPU can reduce the number of decoder layers while increasing the dimensions of its sub-layers to keep the same performance and reduce latency. Figure \ref{fig1} shows the best performing Dynamic-HAT model on the Jetson GPU with a BLEU score of 26.67 has 4 decoder layers. In contrast, the best performing Dynamic-HAT model on the Jetson CPU with a BLEU score of 26.45 has 5 decoder layers. These observations can verify that GPUs search for wide but shallow models – the best Dynamic-HAT model for the GPU has only 4 decoder layers but the largest possible input embedding and hidden dimensions.

\begin{figure}[!htb]
\centerline{\includegraphics[width=9.25cm]{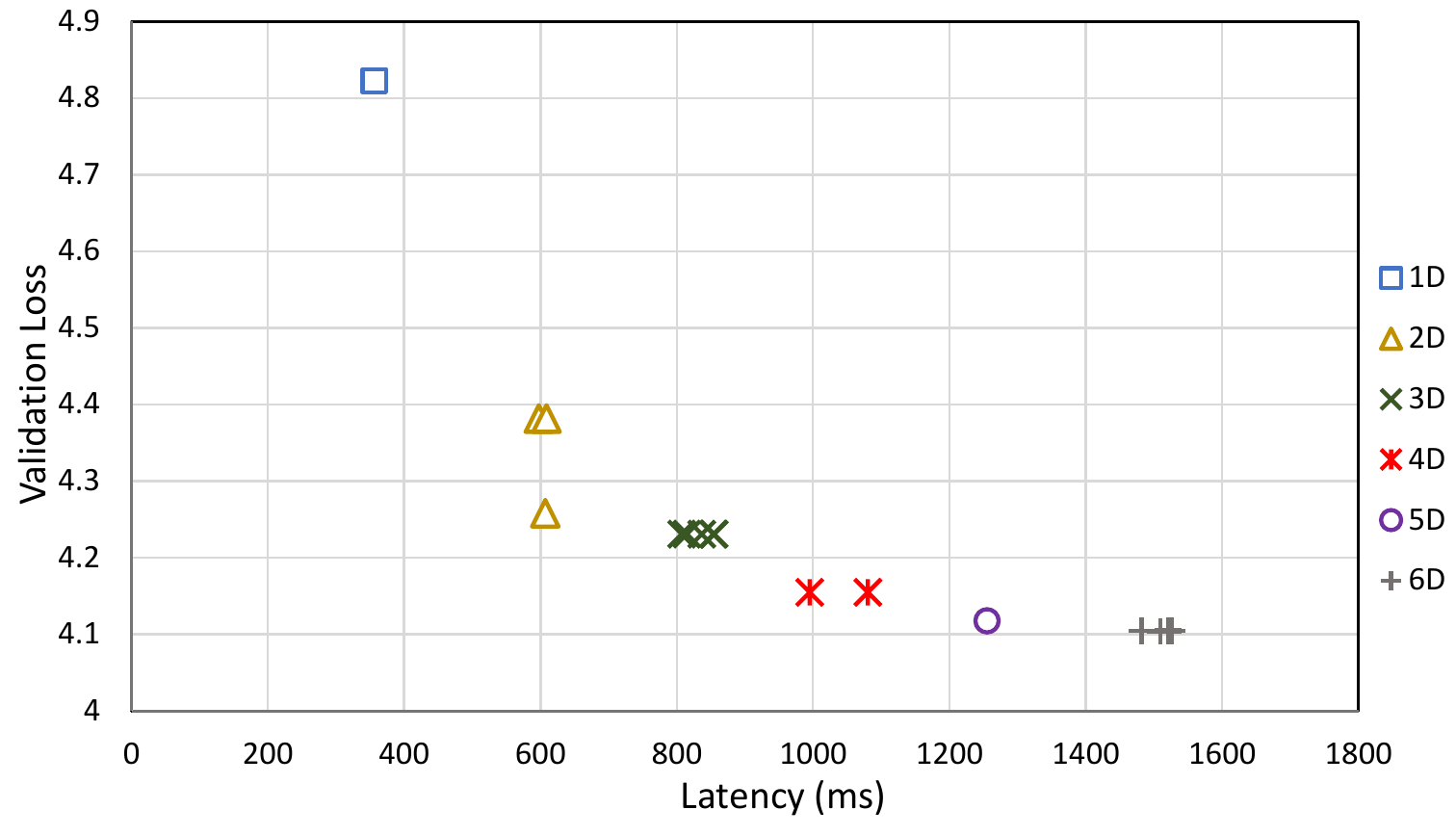}}
\caption{Validation Loss vs experimentally measured Latency on the Jetson Nano GPU, showing how number of decoder layers affect inherited validation loss of the sampled SubTransformer model.}
\label{fig2}
\end{figure}

Table \ref{table2} displays the relationship between the BLEU score and validation loss of inherited SubTransformer configurations and Figure \ref{fig2} shows validation loss of the SubTransformer against measured latency. As the latency constraint increases, validation loss of the searched SubTransformer configuration decreases, as seen in Figure \ref{fig2}, but this change is also not linear. A reduction in validation loss generally improves model performance, leading to higher BLEU scores – this can be observed by comparing Figure \ref{fig1} and Figure \ref{fig2}. In most cases, at latency constraints above 1100ms (for the GPU), 5 or 6 decoder layers are chosen by the evolutionary search – even though the best performing model in terms of BLEU score has 4 decoder layers. As the evolutionary search looks at validation loss, a further decrease in loss would mean that a better performing model has been searched. Although validation loss continues to decrease with the addition of decoder layers and increasing latency constraints, this change is not always reflected in the BLEU score, as seen in Table \ref{table2}.

\begin{table}[]
\centering
\caption {Comparing the switching times of static and dynamic Transformers on the Jetson Nano.}
\begin{tabular}{c|c|c}
\hline
\textbf{Model} & \textbf{Switching Time} & \textbf{Procedure}                                                      \\ \hline
Static HAT \cite{hat}    & minutes to hours        & \begin{tabular}[c]{@{}l@{}}load weights /\\ search / retrain\end{tabular} \\ \hline
Dynamic-HAT    & \textless 1s   & \begin{tabular}[c]{@{}l@{}} sample\end{tabular} \\
\hline
\end{tabular}
\end{table}

\subsection{Performance Improvements Using New Design Space}
The default SuperTransformer design space for HAT is large, as it needs to encompass enough model sizes to span different hardware for a range of latencies. However, we postulate that it may be possible to further optimize SubTransformer sampling and BLEU scores by reducing the dimensions of the design space for a particular platform or product family. As fewer parameters need to be moved around during sampling and inference, the complexity of computation and the number of possible SubTransformer configurations are reduced. We can examine the configurations for the top-5 best performing SubTransformers on the Jetson Nano GPU, and observe which portion of the design space it uses for its optimal SubTransformers. The best performing Jetson Nano models for the GPU, across its entire dynamic range, did not use 640 input embedding dimensions or 1024 hidden dimensions for the FFN layers. The arbitrary encoder-decoder attention sub-layer also only attends to the last one or two encoder layers for optimal SubTransformers on the Jetson GPU. After training for approximately 800 GPU hours on the Nvidia GTX 1080Ti, keeping the same SuperTransformer training setup from HAT \cite{hat}, the following reduced design space is constructed by eliminating the unused parameter choices:

\begin{itemize}
  \item \{512, \sout{640}\} input  embedding  dimensions  for  the  en-coder/decoder
  \item \{\sout{1024}, 2048, 3072\} number of hidden FFN dimensions
  \item \{4, 8\} number  of  attention  heads  in  each  Multi-HeadAttention sub-layer
  \item \{1, 2, 3, 4, 5, 6\} number of decoder layers
  \item \{-1, 1, \sout{2}\} for  arbitrary  encoder-decoder  attention:  -1means  attending  to  last  one  encoder  layer,  1  means  lasttwo encoder layers, 2 means last three encoder layers.
\end{itemize}

\begin{figure}[]
\centerline{\includegraphics[width=9.5cm]{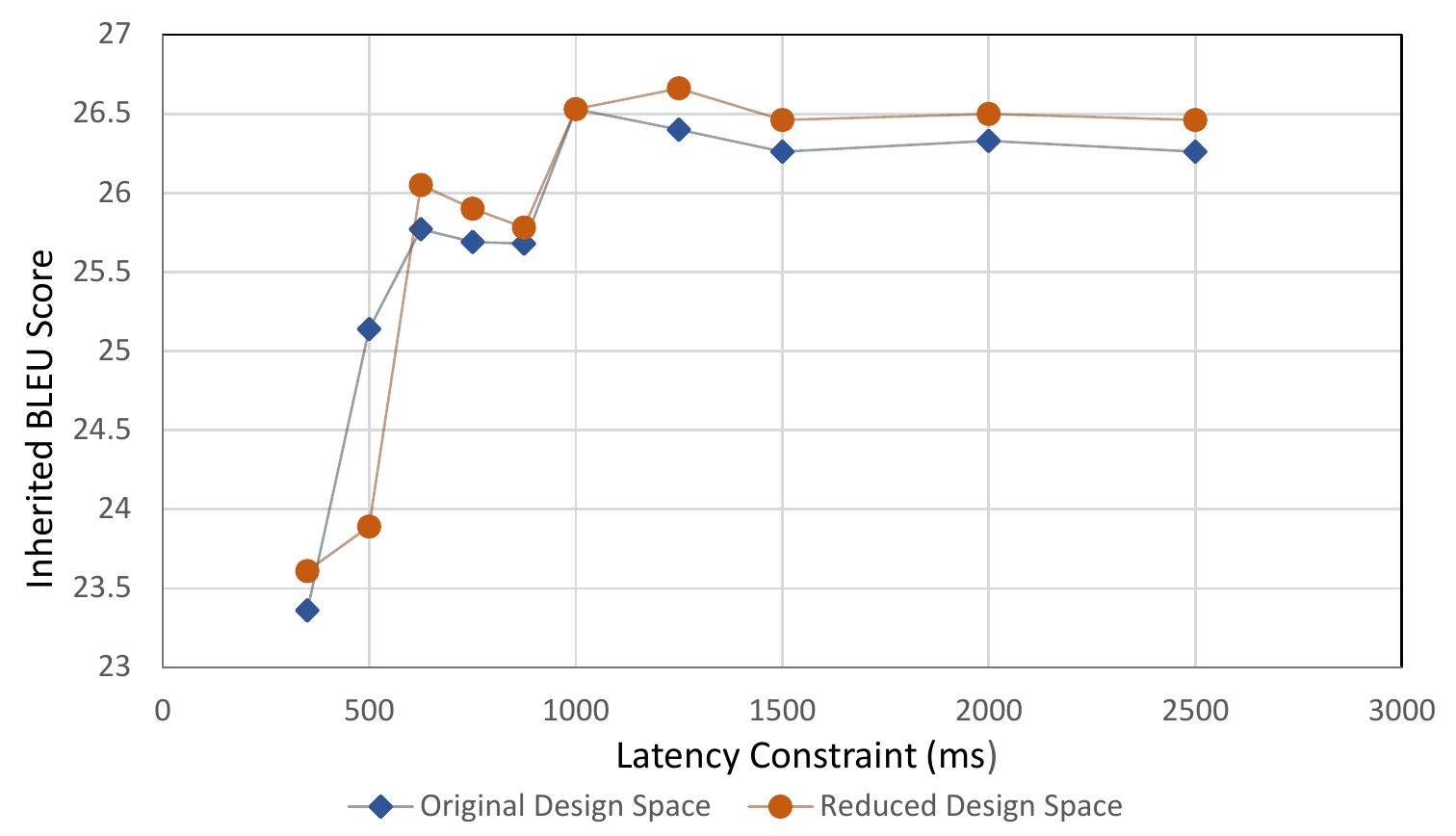}}
\caption{Comparing BLEU scores of operating points for the reduced SuperTransformer design space compared to the original design space.}
\label{fig3}
\end{figure}

The above design space contains $10^{11}$ SubTranformer configurations compared to $10^{15}$ in the original design space. We can observe from Figure \ref{fig3} that the trend of operating points, including sub-optimal points, for the two design spaces follow each other closely. The time to sample a different SubTransformer model from the SuperTransformer, as seen in Table III, is approximately \textless 1s on the Jetson Nano, depending on available resources and model size. The new design space increases the BLEU score of SubTransformers for all latency constraints except 500ms, which appears sub-optimal. All other sub-optimal models from the original design space receive a BLEU score increase of approximately 1\% in the reduced design space.

\section{Conclusion}

Static approaches to optimize a Transformer model for efficient neural machine translation are based on fixed hardware resources. However, it is challenging to consistently meet the performance targets on resource-constrained embedded devices with time-varying availability of hardware resources. Our Dynamic-HAT provides a Transformer model that can dynamically scale to meet latency constraints in different run-time scenarios. A SubTransformer configuration that meets a specified latency constraint can be sampled from the SuperTransformer without retraining in \textless 1s on the Jetson Nano. The proposed Dynamic-HAT uses Inherited SubTransformers. Hence there is a loss in BLEU score $\leqslant$1.5\%. To mitigate the BLEU score loss, we reduce the design space of SuperTransformer. Parameters from the design space that are not used by the Jetson Nano hardware are eliminated, and the SuperTransformer is retrained. The new design space results in a BLEU score increase of approximately 1\% for sub-optimal operating points from the original design space.

\section{Acknowledgement}
This work was supported in part by the Engineering and Physical Sciences Research Council (EPSRC) under Grant EP/S030069/1. Code will be available open-source on https://github.com/UoS-EEC. 



\end{document}